\documentclass[twoside]{article}

\usepackage[accepted]{aistats2023}
\usepackage[round]{natbib}

\usepackage[utf8]{inputenc} 
\usepackage[T1]{fontenc}    
\usepackage{hyperref}       
\usepackage{url}            
\usepackage{booktabs}       
\usepackage{amsfonts}       
\usepackage{nicefrac}       
\usepackage{microtype}      
\usepackage{xcolor}         

\hypersetup{%
    pdfborder = {0 0 0},
    colorlinks,
    citecolor=blue,
    linkcolor=blue,
}

\usepackage{amsmath}
\usepackage{amsthm}
\usepackage{graphicx}
\usepackage{cleveref}

\usepackage{amssymb}
\usepackage{mathtools}

\usepackage[noend]{algpseudocode}
\usepackage{algorithm}

\usepackage{enumitem}
\setlist[description]{leftmargin=*}

\newcommand{\E}{\mathop{\mathbb{E}}} 

\newcommand{\dd}[1]{\mathrm{d}#1}
\newcommand{\AS}{\mathcal{A}_F}
\newcommand{\BS}{\mathcal{B}_F}
\newcommand{\OS}{\mathcal{O}_F}
\newcommand{\ASB}{\mathcal{A}_B}
\newcommand{\BSB}{\mathcal{B}_B}
\newcommand{\OSB}{\mathcal{O}_B}
\newcommand{\ULA}{ULA}
\newcommand{\UHA}{UHA}
\newcommand{\MCD}{MCD}

\newcommand{\LDVI}{LDVI}
\newcommand{\LDVIEM}{\LDVI$_{\mathrm{EM}}$}
\newcommand{\UHAEM}{\UHA$_{\mathrm{EM}}$}
\newcommand{\KL}{\mathrm{KL}}
\newcommand{\ELBO}{\mathrm{ELBO}}
\newcommand{\unnorm}{\bar}
\newcommand{\LP}{\tau_{\mathrm{LP}}}
\newcommand{\LPk}{\tau_{\mathrm{LP},k}}
\newcommand{\std}[1]{\scriptscriptstyle\textcolor{gray}{#1}}

\usepackage{pifont}

\newcommand{\AN}{\mathcal{U}_F}
\newcommand{\BN}{\mathcal{V}_F}
\newcommand{\ANB}{\mathcal{U}_B}
\newcommand{\BNB}{\mathcal{V}_B}

\newcommand{\cstd}{gray}

\newcommand\zg{z_{1:K}}
\newcommand\rhog{\rho_{1:K}}

\usepackage{thmtools} 
\usepackage{thm-restate}

\usepackage{lipsum}
\usepackage{multirow}
\usepackage{wrapfig,booktabs}

\declaretheorem[name=Theorem]{thm}

\newtheorem{theorem}{Theorem}[section]

\newtheorem{lemma}[thm]{Lemma}

\begin{document}

%

%

\twocolumn[

\aistatstitle{Langevin Diffusion Variational Inference}

\aistatsauthor{ Tomas Geffner \And Justin Domke}

\aistatsaddress{ University of Massachusetts, Amherst\\\texttt{tgeffner@cs.umass.edu} \And  University of Massachusetts, Amherst\\\texttt{domke@cs.umass.edu}} ]

\begin{abstract}
Many methods that build powerful variational distributions based on unadjusted Langevin transitions exist. Most of these were developed using a wide range of different approaches and techniques. Unfortunately, the lack of a unified analysis and derivation makes developing new methods and reasoning about existing ones a challenging task. We address this giving a single analysis that unifies and generalizes these existing techniques. The main idea is to augment the target and variational by numerically simulating the underdamped Langevin diffusion process and its time reversal. The benefits of this approach are twofold: it provides a unified formulation for many existing methods, and it simplifies the development of new ones. In fact, using our formulation we propose a new method that combines the strengths of previously existing algorithms; it uses underdamped Langevin transitions and powerful augmentations parameterized by a score network. Our empirical evaluation shows that our proposed method consistently outperforms relevant baselines in a wide range of tasks.
\end{abstract}

\section{INTRODUCTION}

Several recent work attempts to build powerful variational distributions using unadjusted Hamiltonian Monte Carlo (HMC) transition kernels \citep{salimans2015markov, wolf2016variational, caterini2018hamiltonian, wu2020stochastic, thin2021monte, dais, UHA, chen2022bayesian}. In principle, one would like to use the last sample marginal of the HMC chain as variational distribution. Since this marginalization is typically intractable, these methods use auxiliary variables \citep{agakov2004auxiliary}; they build an augmented variational distribution that includes all samples generated in the chain, an augmented target, and perform variational inference (VI) on these augmented distributions. Training proceeds by maximizing the ELBO using unbiased reparameterization gradients, made possible by using \textit{uncorrected} transitions.

One such method is Unadjusted Langevin Annealing (\ULA) \citep{wu2020stochastic, thin2021monte}, which can be seen as an approximation of Annealed Importance Sampling \citep{neal2001ais, jarzynski1997nonequilibrium}. The method builds a sequence of densities that gradually bridge an initial approximation to the target, and augments the variational distribution and target using uncorrected overdamped Langevin kernels targeting each of these bridging densities.

While \ULA\ has shown good performance, it has two limitations: It is based on overdamped Langevin dynamics, which are known to suffer from random walk behavior \cite[\S5.2]{neal2011mcmc}, and it augments the target using an approximation of the Annealed Importance Sampling augmentation, which is known to be suboptimal \citep{doucet2006sequential}. These two limitations were addressed independently. Uncorrected Hamiltonian Annealing (\UHA) \citep{UHA,dais} extends \ULA\ to use \emph{underdamped} Langevin transitions, known to improve convergence over the overdamped variant \citep{cheng2018underdamped}. Meanwhile, Monte Carlo Diffusion (\MCD) \citep{doucet2022annealed} extends \ULA\ to use better augmentations for the target. Both of these lead to significant performance improvements over \ULA, albeit through orthogonal enhancements.

These methods were developed using different approaches: While \UHA\ was developed as a differentiable approximation to Annealed Importance Sampling with underdamped Langevin transitions, \MCD\ was developed by numerically simulating the overdamped Langevin diffusion and its time reversal, approximating intractable terms with a score network \citep{song2019generative}. The fact that these methods have different derivations and are based on different techniques makes it difficult to reason about their benefits, drawbacks, and the connections between them. It also means it is not obvious how to combine both of their benefits.

This paper introduces a formulation for Langevin-based VI that encompasses previously proposed methods. This formulation can be seen as a generalization of \MCD\ \citep{doucet2022annealed} that uses underdamped Langevin dynamics, instead of the overdamped variant. Like \MCD\ \citep{doucet2022annealed}, our approach is based on the analysis of continuous time processes. Its main components are the underdamped Langevin diffusion process and its time reversal, which are numerically simulated to derive the augmentations for the variational approximation and target. We introduce our approach for Langevin-based VI in \cref{sec:ldvi}.

Our method is compatible with multiple numerical simulation schemes, with different choices leading to different algorithms. \Cref{sec:discr} introduces a simulation scheme based on splitting methods \citep{bou2010long, melchionna2007design}. We show that this specific scheme can be used to recover \ULA, \MCD\ and \UHA, providing a unified view for all of them, and shedding light on the connections between them, their benefits and limitations.

Additionally, our formulation facilitates the development of new methods. We use it to propose Langevin Diffusion VI (\LDVI), a novel method that combines the best of \UHA\ and \MCD: it uses powerful and improved augmentations for the target, like \MCD, while enjoying the benefits of underdamped Langevin transitions, like \UHA. We evaluate \LDVI\ empirically in \cref{sec:exps}, showing that it outperforms \ULA, \UHA\ and \MCD\ in a range of inference tasks.

Finally, we explore the importance of the numerical simulation scheme. In \cref{sec:naive} we observe that one can also develop methods using a Euler-Maruyama type discretization scheme. Our experimental results therein show that the simulation method used plays a crucial role in the algorithms' performance, suggesting a possible direction to explore to further improve these methods.

\section{PRELIMINARIES} \label{sec:prels}

\paragraph{Variational Inference.} VI approximates a target distribution $p(z) = \unnorm p(z) / Z$ (known up to the normalizing constant $Z$) with a simpler distribution $q$. It works by finding the parameters of $q$ that maximize the evidence lower bound
\begin{equation}
\ELBO(q(z) \Vert \unnorm p(z)) = \E_{q(z)} \log \frac{\unnorm p(z)}{q(z)}. \label{eq:ELBO}
\end{equation}
Noting that $\log Z = \ELBO(q(z) \Vert \unnorm p(z)) + \KL(q(z) \Vert p(z))$, it can be seen that maximizing the ELBO is equivalent to minimizing the KL-divergence from the approximation $q(z)$ to the target $p(z)$.

\paragraph{MCMC-VI.} Many methods have been developed to use MCMC to build powerful variational approximations. Ideally, one would use the last sample marginal of an MCMC chain as the approximating distribution. However, since computing this marginal is typically intractable, most methods are based on augmentations \citep{agakov2004auxiliary} and variants of Annealed Importance Sampling \citep{wu2020stochastic, thin2021monte, UHA, dais, doucet2022annealed}. They define a sequence of unnormalized densities $\unnorm \pi_k(z) = q(z)^{1 - \beta_k} \unnorm p(z)^{\beta_{k}}$, for $k=1, \hdots, K-1$ and $0 < \beta_1 < \hdots < \beta_{K-1} < 1$, forward transitions $F_k(z_{k+1} \vert z_k)$ that (approximately) leave $\pi_k$ invariant, backward transitions $B_k(z_k\vert z_{k+1})$, and build the augmented target and variational distribution as
\begin{equation}
\begin{split} \label{eq:augmentations} \textstyle
q(\zg) & = q(z_1) \prod_{k=1}^{K-1} F_k(z_{k+1}\vert z_k)\\
\unnorm p(\zg) & = \unnorm p(z_K) \prod_{k=1}^{K-1} B_k(z_{k}\vert z_{k+1}).
\end{split}
\end{equation}
Then, one attempts to tune the forward and backward transitions to maximize the ELBO between these augmented distributions, equivalent to minimizing the KL divergence between them. The chain rule for the KL-divergence \citep{cover1999elements} then guarantees $\KL(q(z_K)\Vert p(z_K)) \leq \KL(q(\zg) \Vert p(\zg))$, justifying the use of the marginal of $q(\zg)$ over $z_K$ to approximate the original target distribution.

While augmentations bypass intractable marginalizations, they introduce additional looseness in that $\ELBO(q(\zg) \Vert p(\zg)) \leq \ELBO(q(z_K)\Vert p(z_K))$. For a given set of forward transitions $F_k$, this inequality can in principle be made tight by using the optimal backward transitions \citep{doucet2006sequential}
\begin{equation} \label{eq:optB}
B_k(z_k \vert z_{k+1}) = F_k(z_{k+1} \vert z_k) \frac{q(z_k)}{q(z_{k+1})}.
\end{equation}
In practice, however, the marginal densities $q(z_k)$ are not exactly known, so algorithms must use other choices for $B_k$. There are two desiderata: the ratio $\nicefrac{B_k}{F_k}$ must be tractable (required to get a tractable expression for the ELBO between the augmented distributions), and the transitions should be differentiable (not strictly needed, but desirable, as it allows the use of reparameterization gradients to tune all parameters). Most recent methods were developed with these two properties in mind \citep{salimans2015markov, wolf2016variational, wu2020stochastic, thin2021monte, UHA, dais, doucet2022annealed, jankowiak2021surrogate}. For instance, \ULA\ uses unadjusted overdamped Langevin kernels for both $F_k$ and $B_k$, and \UHA\ extends it to use underdamped Langevin kernels. For the latter, the distributions from \cref{eq:augmentations} are further augmented to include momentum variables $\rho$, leading to
\begin{equation}\begin{split} \label{eq:augmomtransitions}
q(\zg, \rhog) &= q(z_1, \rho_1) \prod_{k=1}^{K-1} F_k(z_{k+1}, \rho_{k+1}\vert z_k, \rho_k)\\
\unnorm p(\zg, \rhog) &= \unnorm p(z_K, \rho_K) \prod_{k=1}^{K-1} B_k(z_{k}, \rho_k\vert z_{k+1}, \rho_{k+1}),\\
\end{split}\end{equation}
and to the augmented ELBO
\begin{multline} \label{eq:augelbo}
\ELBO(q(\zg, \rhog) \Vert \unnorm p(\zg, \rhog)) = \\ \E_{q} \left[ \log \frac{\unnorm p(z_K, \rho_K)}{q(z_1, \rho_1)} + \sum_{k=1}^{K-1} \log \frac{F_k(z_{k+1}, \rho_{k+1}\vert z_k, \rho_k)}{B_k(z_{k}, \rho_k\vert z_{k+1}, \rho_{k+1})}\right].
\end{multline}

\section{LANGEVIN DIFFUSION VARIATIONAL INFERENCE} \label{sec:ldvi}

This section introduces our approach for Langevin-based VI. It provides a way to build the augmented distributions from \cref{eq:augmomtransitions}. Its main components are (1) the underdamped Langevin diffusion process and its time reversal, (2) a numerical simulation scheme to approximately simulate these processes, and (3) a score network \citep{doucet2022annealed, song2019generative} used to approximate intractable terms in the time-reversed process. Together, these produce the forward and backward transitions $F_k$ and $B_k$ with a tractable ratio. Since our approach is compatible with many simulation schemes, we first introduce it in a general way, and present a specific simulation scheme in \cref{sec:discr}.

\subsection{Langevin Diffusion} \label{sec:langevin}

This sub-section introduces the Langevin diffusion process and its time reversal, which will be used to derive the forward and backward transitions in the following sections. Let $\pi^t(z)$ be a sequence of densities bridging from the starting distribution for $t=0$ to the target for $t=T$. That is, $\pi^0(z)=q(z)$ and $\pi^T(z)=p(z)$. The Langevin diffusion process is characterized by the following stochastic differential equation (SDE):
\begin{equation} \label{eq:fwSDE}
\begin{split}
\dd{z^t} & = \rho^t \dd{t}\\
\dd{\rho^t} & = \left[\nabla \log \pi^t(z^t) - \gamma \rho^t \right] \dd{t} + \sqrt{2\gamma} \, \dd{w^t},
\end{split}
\end{equation}
where $t\in[0,T]$, $w^t$ is a standard Wiener process, $\gamma > 0$ is a friction coefficient, and $(z^0, \rho^0) \sim q(z^0, \rho^0)$. The forward transitions $F_k$ will be derived by simulating this process. The motivation behind the use of this process comes from its good convergence properties. Intuitively, evolving \cref{eq:fwSDE} yields values for $(z^t, \rho^t)$ that tend to be close to $\pi^{t}(z) \mathcal{N}(\rho \vert 0,I)$. Thus, one may hope that the marginal density of the process at time $T$ is close to $p(z) \mathcal{N}(\rho \vert 0,I)$, meaning the distribution of the final value $z^T$ may be close to the target of interest.

The backward transitions $B_k$, on the other hand, will be derived by simulating the time-reversed SDE corresponding to \cref{eq:fwSDE}. Defining $y^t = z^{T-t}$ and $\lambda^t = \rho^{T-t}$, this time-reversed process is characterized by (obtained using results for time-reversed diffusions \citep{anderson1982reverse, haussmann1986time},  see \cref{app:reverseSDE})
\begin{equation} \label{eq:bwSDE}
\begin{split}
    \dd{y^t} = & -\lambda^t \dd t\\
    \dd{\lambda^t} = & [ \gamma \lambda^t - \nabla \log \pi^{T-t}(y^t) + \\
    & \,\,2\gamma \nabla_\lambda \log q^{T-t}(y^t, \lambda^t) ] \dd t + \sqrt{2\gamma} \, \dd{w^t},
\end{split}
\end{equation}
where $q^{t}$ is the marginal of the forward process at time $t$. Formally, this process is initialized with $(y^0, \lambda^0) \sim q^T(y^0, \lambda^0)$. However, in what follows, where we use it to define the backward transitions to augment the target, it will be initialized as $(y^0, \lambda^0) \sim p(y^0, \lambda^0)$. The motivation for using the reverse time SDE from \cref{eq:bwSDE} is that, under exact simulation, it yields the optimal backward transitions from \cref{eq:optB} (i.e. no additional looseness in the augmented ELBO).

\subsection{Transitions via SDE Simulation}

The forward and backward transitions will be obtained by simulating the forward and time-reversed processes for a fixed period of time $\delta = \nicefrac{T}{K}$. If we could simulate the above SDEs \textit{exactly}, then
\begin{itemize}[leftmargin=*] \setlength\itemsep{0pt}
\vspace{-0.1cm}
\item The forward transition $F_k(z_{k+1}, \rho_{k+1} \vert z_k, \rho_k)$ would be obtained by simulating the forward process from time $t=k\delta$ up to time $t=(k+1)\delta$, starting from the initial values $(z, \rho) = (z_k, \rho_k)$,
\item The backward transition $B_k(z_{k}, \rho_{k} \vert z_{k+1}, \rho_{k+1})$ would be obtained by simulating the reverse-time SDE from time $t=(K-k)\delta$ up to time $t=(K-k-1)\delta$, starting from the initial values $(y, \lambda) = (z_{k+1}, \rho_{k+1})$.
\vspace{-0.1cm}
\end{itemize}
It can be shown that these backward transitions are optimal. That is, if one could simulate \cref{eq:fwSDE,,eq:bwSDE} exactly to get the forward and backward transitions defined above, the resulting augmentations would be tight in the sense that the augmented ELBO from \cref{eq:augelbo} would have no additional looseness compared to an ELBO defined between the last sample marginals $q_K(z_K,\rho_K)$ and $\bar{p}(z_K,\rho_K)$.

Unfortunately, these transitions are intractable for two reasons. First, the forward marginal density $q^{t}$ that appears in the reverse SDE is unknown. Second, it is intractable to exactly simulate or evaluate either of the above SDEs.

\paragraph{Approximating $\mathbf{\nabla_\lambda \mbox{\textbf{log}}\, q^{T-t}(y^t, \lambda^t)}$.}

The first source of intractability of the optimal transitions is the score term $\nabla_\lambda \log q^{T-t}(y^t, \lambda^t)$, which is typically unavailable. Inspired by the fact that $q^{T-t}(y, \lambda)$ is expected to be close to $\pi^{T-t}(y)\mathcal{N}(\lambda\vert 0, I)$, we propose to approximate this term as
\begin{equation} \label{eq:approx_grad}
\nabla_\lambda \log q^{T-t}(y^t, \lambda^t) \approx -\lambda^t + s(T-t, y^t, \lambda^t),
\end{equation}
where $s: \mathbb{R} \times \mathbb{R}^D \times \mathbb{R}^D \rightarrow \mathbb{R}^D$ is some learnable function approximator. Following recent work \citep{song2020score, song2019generative, sohl2015deep, ho2020denoising, doucet2022annealed}, we use a neural network, typically referred to as score network, which is trained with the other parameters to maximize the ELBO. The intuition behind our approximation in \cref{eq:approx_grad} comes from considering scenarios where the forward transitions mix fast. In such cases $q^{T-t}(y^t, \lambda^t)$ will be close to $\pi^{T-t}(y^t) \mathcal{N}(\lambda^t \vert 0, I)$, and thus the approximation $\nabla_\lambda \log q^{T-t}(y^t, \lambda^t) \approx -\lambda^t$ should work well. (In fact, as we show in \cref{sec:recovering_methods}, several well-known methods are recovered by removing the score network; that is, fixing $s(t, y, \lambda) = 0$.)

\paragraph{Transitions via numerical simulation.} The second source of intractability is that it is rarely possible to simulate the forward and reverse SDEs \textit{exactly}. Thus, we use a numerical simulation scheme to \textit{approximately} simulate them.
The requirements for the simulation scheme are (1) it must yield transitions with a tractable ratio, and (2) it must be differentiable, in order to allow unbiased reparameterization gradients \citep{doublystochastic_titsias, vaes_welling, rezende2014stochastic}. \Cref{sec:discr} presents a scheme that satisfies these.

\subsection{Framework for Langevin-based VI} \label{sec:framework}

Our formulation for Langevin-based VI is based on the transitions described above. To get a specific instance, several choices are required:
\begin{itemize}[leftmargin=*] \setlength\itemsep{0pt}
\vspace{-0.1cm}
\item A momentum augmented target $\unnorm p(z_K, \rho_K) = \unnorm p(z_K) p(\rho_K \vert z_K)$ that retains original target $\unnorm p(z)$ as marginal, often defined as $\unnorm p(z_K) \mathcal{N}(\rho_K \vert 0, I)$,
\item A momentum augmented initial approximation $q(z_1, \rho_1)$, often defined as $q(z_1) \mathcal{N}(\rho_1 \vert 0, I)$,
\item A score network $s(t, z, \rho)$ to approximate intractable term involving $q^t(z, \rho)$,
\item Forward and backward transitions $F_k$ and $B_k$ with a tractable ratio, obtained by numerically simulating the forward and reverse SDEs from \cref{eq:fwSDE,,eq:bwSDE}.
\end{itemize}
For specific choices for these components, we can compute
\begin{equation} \label{eq:ratio}
\frac{\unnorm p(\zg, \rhog)}{q(\zg, \rhog)} = \frac{\unnorm p(z_K, \rho_K)}{q(z_1, \rho_1)} \prod_{k=1}^{K-1} \frac{B_k(z_{k}, \rho_{k} \vert z_{k+1}, \rho_{k+1})}{F_k(z_{k+1}, \rho_{k+1} \vert z_k, \rho_k)},
\end{equation}
required to estimate and optimize the ELBO from \cref{eq:augelbo}.

\section{NUMERICAL SIMULATION SCHEME} \label{sec:discr}

This section introduces two numerical simulation schemes, one for the forward SDE and one for the time-reversed SDE, which yield transitions with a tractable ratio. We begin by giving explicit algorithmic representations for these transitions and an expression for their ratio (\cref{sec:algtransitions}). We then explain how our formulation for Langevin-based VI with these transitions can be used to recover several existing methods, including \ULA, \MCD\ and \UHA\ (\cref{sec:recovering_methods}), and also how it can be used to derive new methods (\cref{sec:newmethod}).

\subsection{Forward and Backward Transitions} \label{sec:algtransitions}

The forward transitions used to approximately simulate to forward SDE are shown in \cref{alg:Fldvi}. They consist of two steps: (partial) momentum resampling from some distribution $m_F$ (see \cref{sec:dertransitions}), followed by a single leapfrog integrator step typically used to simulate Hamiltonian dynamics \citep{neal2011mcmc, betancourt2017conceptual} (denoted by $\LP$ in \cref{alg:Fldvi}, which consists on sequential deterministic updates to the variables $\rho$, $z$, and $\rho$). As explained in \cref{sec:dertransitions}, these transitions are derived by simulating the forward SDE from \cref{eq:fwSDE} using splitting methods \citep{bou2010long, melchionna2007design}.


\begin{algorithm}[t]
\caption{Forward transition $F_k(z_{k+1}, \rho_{k+1} \vert z_{k}, \rho_{k})$}
\label{alg:Fldvi}
\begin{algorithmic}
\Require $z_k$, $\rho_k$, step-size $\delta$
\State Re-sample momentum $\rho_k' \sim m_F(\rho_k'\vert \rho_k, \gamma, \delta)$\\
\hspace{-0.21cm}$\left.\begin{array}{l} \vspace{0.05cm}
\mbox{Update} \, \rho_k'' = \rho_k' + \frac{\delta}{2} \nabla \log \pi^{k\delta}(z_k)\\\vspace{0.05cm}
\mbox{Update} \, z_{k+1} = z_k + \delta \rho_k''\\
\mbox{Update} \, \rho_{k+1} = \rho_k'' + \frac{\delta}{2} \nabla \log \pi^{k\delta}(z_{k+1})
\end{array}\right\}
\left . \begin{array}{l}
\mbox{Leapfrog step}\\
\LP(z_k, \rho_k')
\end{array}\right .$
\State \Return $(z_{k+1}, \rho_{k+1})$
\end{algorithmic}
\end{algorithm}

The backward transitions used to approximately simulate the time-reversed SDE are shown in \cref{alg:Bldvi}. They also consist of two steps: the \textit{inverse} of a single leapfrog integrator step used to simulate Hamiltonian dynamics, followed by a (partial) momentum resampling from some distribution $m_B$. We include their derivation and details for the momentum resampling distribution $m_B$ in \cref{sec:dertransitions}.


\begin{algorithm}[t]
\caption{Backward transition $B_k(z_{k}, \rho_{k} \vert z_{k+1}, \rho_{k+1})$}
\label{alg:Bldvi}
\begin{algorithmic}
\Require $z_{k+1}$, $\rho_{k+1}$, step-size $\delta$
\State
\hspace{-0.21cm}$\left.\begin{array}{l} \vspace{0.05cm}
\mbox{Update} \, \rho_k'' = \rho_{k+1} - \frac{\delta}{2} \nabla \log \pi_k(z_k)\\\vspace{0.05cm}
\mbox{Update} \, z_{k} = z_{k+1} - \delta \rho_k''\\
\mbox{Update} \, \rho_{k}' = \rho_k'' - \frac{\delta}{2} \nabla \log \pi_k(z_{k+1})
\end{array}\right\}
\left . \begin{array}{l}
\mbox{Inverse leapfrog}\\
\LP^{-1}(z_{k+1}, \rho_{k+1})
\end{array}\right .$
\State Re-sample momentum $\rho_k \sim m_B(\rho_k\vert \rho_k', z_k, \gamma, \delta)$
\State \Return $(z_{k}, \rho_{k})$
\end{algorithmic}
\end{algorithm}

In order to use these transitions for Langevin-based VI, we need an expression for their ratio. This is given in \cref{lemma:ratio}, proved in \cref{app:proofs}.
\begin{lemma} \label{lemma:ratio}
Let $F_k(z_{k+1}, \rho_{k+1} \vert z_{k}, \rho_{k})$ and $B_k(z_{k}, \rho_{k} \vert z_{k+1}, \rho_{k+1})$ be the transitions defined in \cref{alg:Fldvi,,alg:Bldvi}, $m_F$ and $m_B$ the momentum resampling distributions used in these transitions, $\delta$ the discretization step-size, and $\gamma > 0$ the damping coefficient. Then, 
\begin{equation}
\frac{B_k(z_{k}, \rho_{k} \vert z_{k+1}, \rho_{k+1})}{F_k(z_{k+1}, \rho_{k+1} \vert z_{k}, \rho_{k})} = \frac{m_B(\rho_k\vert \rho_k', z_k, \gamma, \delta)}{m_F(\rho_k'\vert \rho_k, \gamma, \delta)},
\end{equation}
where $\rho_k'$ is as defined in \cref{alg:Fldvi,,alg:Bldvi}, given by $(z_{k}, \rho_{k}') = \LP^{-1}(z_{k+1}, \rho_{k+1})$.
\end{lemma}

Using the transitions from \cref{alg:Fldvi,,alg:Bldvi} and their ratio given in \cref{lemma:ratio} we can get an exact expression for the augmented ELBO from \cref{eq:augelbo}. While computing this augmented ELBO exactly is typically intractable, an unbiased estimate can be obtained using a sample from $q(\zg, \rhog)$, as shown in \cref{alg:gen_aug_elbo}.

\begin{algorithm}[ht]
\caption{Generating the augmented ELBO (\cref{eq:augelbo}).}
\label{alg:gen_aug_elbo}
\begin{algorithmic}
\State Sample $(z_1, \rho_1) \sim q(z_1, \rho_1)$.
\State Initialize estimator as $\mathcal{L} \leftarrow -\log q(z_1, \rho_1)$.
\For{$k = 1, 2, \cdots , K-1$}
	\State Run $F_k$ (alg.~\ref{alg:Fldvi}) on $(z_k, \rho_k)$, store $\rho'_k, z_{k+1}, \rho_{k+1}$.
	\State Update $\mathcal{L} \leftarrow \mathcal{L} + \log \frac{m_B(\rho_k\vert \rho_k', z_k, \gamma, \delta)}{m_F(\rho_k'\vert \rho_k, \gamma, \delta)}$.
\EndFor
\State Update $\mathcal{L} \leftarrow \mathcal{L} + \log \unnorm p(z_K, \rho_K)$.
\State \Return $\mathcal{L}$
\end{algorithmic}
\end{algorithm}

\subsubsection{Derivation of Forward and Backward Transitions} \label{sec:dertransitions}

We now show the derivation for the forward and backward transitions using splitting methods \citep{bou2010long, melchionna2007design}, which have been observed to work well for Langevin processes \citep{leimkuhler2013rational, monmarche2021high}. Simply put, splitting methods split an SDE into multiple simpler components, simulate each component for a time-step of size $\delta$, and then combine the solutions sequentially to build the $\delta$-sized step for the original SDE.

\paragraph{Forward transitions.} These are obtained by approximately simulating the forward SDE using a splitting method. Following Monmarche \citep{monmarche2021high}, we split the SDE in three components, $\AS$, $\BS$ and $\OS$,\footnote{A similar split was used in the context of generative modeling by Dockhorn et al. \citep{dockhorn2021score}, albeit for a different (simpler) diffusion which targets a Gaussian using a different definition for the bridging densities, as typically done with diffusion models \citep{sohl2015deep}.}
\begin{equation*}
\small
\underbrace{\left[\begin{array}{c}
\dd z^t\\
\dd \rho^t
\end{array}\right] = 
\left[\begin{array}{c}
\rho^t \dd t \\
0
\end{array}\right]}_{\AS}, \quad 
\underbrace{\left[\begin{array}{c}
\dd z^t\\
\dd \rho^t
\end{array}\right] = 
\left[\begin{array}{c}
0\\
\nabla \log \pi^t(z^t) \dd t
\end{array}\right]}_{\BS},
\end{equation*}
\begin{equation*}
\small
\underbrace{\left[\begin{array}{c}
\dd z^t\\
\dd \rho^t
\end{array}\right] = 
\left[\begin{array}{c}
0\\
-\gamma \rho^t \dd t + \sqrt{2\gamma} \dd w^t
\end{array}\right]}_{\OS},
\end{equation*}
each one simpler than the original SDE, and then build the forward transition by sequentially composing the simulations for components $\OS\BS\AS\BS$. The final forward transition shown in \cref{alg:Fldvi} can be obtained by noting that each of the individual components can be simulated with the following strategies:
\vspace{-0.2cm}
\begin{description}
	\item[Simulating $\AS$.] This can be done exactly. Given initial values $(z^{t_0}, \rho^{t_0})$ at time $t_0$, simulating $\AS$ for a time $\delta$ results in $(z^{t_0+\delta}, \rho^{t_0+\delta}) = (z^{t_0} + \delta \rho^{t_0}, \rho^{t_0})$.
	\item[Simulating $\BS$.] Given initial values $(z^{t_0}, \rho^{t_0})$ at time $t_0$, and using that $\pi^{t_0} \approx \pi^{t_0+\delta}$ for small $\delta$, simulating $\BS$ for a time $\delta$ results in $(z^{t_0+\delta}, \rho^{t_0+\delta}) = (z^{t_0}, \rho^{t_0} + \delta \nabla \log \pi^{t_0}(z^{t_0}))$.
	\item[Simulating $\OS$.] This can be done exactly, as $\OS$ corresponds to an Ornstein–Uhlenbeck process. Given an initial value of $\rho^{t_0}$ at time $t_0$, simulating $\OS$ for a time $\delta$ gives $\rho^{t_0+\delta} \sim \mathcal{N}(\rho^{t_0+\delta} \vert \eta \rho^{t_0}, (1 - \eta^2) I)$, where $\eta = \exp(-\gamma \delta)$. However, as we will see next, exact simulation for the corresponding component of the reverse SDE is not possible. Thus, it may be useful to simulate $\OS$ approximately as well, using the Euler-Maruyama scheme \citep{maruyama1955continuous, bayram2018numerical}, which gives $\rho^{t_0+\delta} \sim \mathcal{N}(\rho^{t_0+\delta} \vert \rho^{t_0}(1 - \gamma\delta), 2\gamma\delta I)$. We use $m_F$ to denote generically the momentum resampling distribution used, which could be any of the ones just described.
\end{description}
In summary, simulating $\OS$ yields the momentum resampling step, while composing the simulations for $\BS\AS\BS$ yields the leapfrog integration step (note that since $\BS$ is simulated twice, it is done with a step-size of $\nicefrac{\delta}{2}$.)

\paragraph{Backward transitions.} Like for the forward transitions, these are derived by splitting the reverse SDE in three components, $\ASB$, $\BSB$ and $\OSB$ (using our approximation for the score term),
\begin{equation*}
\small
\underbrace{\left[\begin{array}{c}
\dd y^t\\
\dd \lambda^t
\end{array}\right] = 
\left[\begin{array}{c}
-\lambda^t \dd t\\
0
\end{array}\right]}_{\ASB}, \,\, 
\underbrace{\left[\begin{array}{c}
\dd y^t\\
\dd \lambda^t
\end{array}\right] = 
\left[\begin{array}{c}
0\\
-\nabla \log \pi^{T-t}(y^t) \dd t
\end{array}\right]}_{\BSB},
\end{equation*}
\begin{equation*}
\small
\underbrace{\left[\begin{array}{c}
\dd y^t\\
\dd \lambda^t
\end{array}\right] = 
\left[\begin{array}{c}
0\\
-\gamma \lambda^t \dd t + 2\gamma s(T-t, y^t, \lambda^t) \dd t + \sqrt{2\gamma} \dd w^t
\end{array}\right]}_{\OSB},
\end{equation*}
Then, we construct the backward transition by sequentially composing the simulations for components $\BSB\ASB\BSB\OSB$, where the sequence $\BSB\ASB\BSB$ yields the inverse of the leapfrog integrator step, and $\OSB$ yields the momentum resampling step. The derivation follows the one for the forward transitions closely, with one main difference: simulating component $\OSB$ has an additional difficulty, due to the presence of the term involving the score network $s$. While in general $\OSB$ cannot be simulated exactly (unless we fix $s=0$), it can be done approximately using the Euler-Maruyama method, which results in the momentum resampling distribution $m_B(\lambda_{t+\delta} \vert \lambda_t, y_t, \gamma, \delta) = \mathcal{N}(\lambda_{t+\delta}\vert \lambda_t (1 - \gamma\delta) + 2 \gamma \delta s(T-t, y_t, \lambda_t), 2\gamma\delta I)$. We give further details in \cref{app:derivingB}.

\subsection{Recovering Known Methods} \label{sec:recovering_methods}

As mentioned previously, \ULA, \MCD\ and \UHA\ were originally derived using different techniques and approaches. Some of these methods use overdamped Langevin dynamics, while others use the underdamped variant; some were derived as approximations of Annealed Importance Sampling, while others emerged from an analysis of continuous time diffusion processes. This section's main purpose is to show that all of these methods can be derived in a unified way using the formulation for Langevin-based VI from \cref{sec:framework} with the numerical simulation schemes introduced above. We begin by briefly giving details about \ULA, \MCD\ and \UHA\ (including their different derivations), followed by an explanation of how these methods can be recovered with our approach.

\textbf{\ULA} \citep{wu2020stochastic, thin2021monte} works directly with unadjusted overdamped Langevin kernels (i.e. no momentum variables), defining the forward transitions as
\begin{equation}
F_k(z_{k+1} \vert z_k) = \mathcal{N}(z_{k+1} \vert z_k + \delta \nabla \log \pi_k(z_k), 2\delta I).
\end{equation}
Then, using the fact that $F_k(z_{k+1} \vert z_k)$ is approximately reversible with respect to $\pi_k$ when the step-size $\delta$ is small, it defines the backward transitions as $B_k(z_k \vert z_{k+1}) = F_k(z_k\vert z_{k+1})$. The ratio between these transitions, and thus the augmented ELBO, are straightforward to compute (see \cref{app:proofs}). Broadly speaking, the method can be seen as a differentiable approximation to Annealed Importance Sampling with overdamped Langevin transitions.

\begin{theorem}\label{thm:ula}
\ULA\ is recovered by the formulation from \cref{sec:framework} with $\unnorm p(z_K, \rho_k) = \unnorm p(z_K) \mathcal{N}(\rho_K\vert 0, I)$, $q(z_1, \rho_1) = q(z_1) \mathcal{N}(\rho_1\vert 0, I)$, $s(t, z, \rho) = 0$, and the transitions from \cref{alg:Fldvi,,alg:Bldvi} with exact momentum resampling (possible due to removing the score network) with $\eta = 0$ (high friction limit).
\end{theorem}

\textbf{\MCD} \citep{doucet2022annealed} was developed by studying the overdamped Langevin diffusion process, given by
\begin{equation} \label{eq:odl}
\dd z^t = \nabla \log \pi^t(z^t) \dd t + \sqrt{2} \, \dd w^t.
\end{equation}
It uses unadjusted overdamped Langevin kernels for the forward transitions (i.e. simulating \cref{eq:odl} with the Euler-Maruyama scheme), and uses backward transitions derived by simulating the reverse-time diffusion corresponding to \cref{eq:odl}, also with the Euler-Maruyama scheme, using a score network to approximate intractable terms.

\begin{theorem}\label{thm:mcd}
\MCD\ is recovered by the formulation from \cref{sec:framework} with $s(t, z, \rho) = \tilde s(t, z)$, $\unnorm p(z_K, \rho_k) = \unnorm p(z_K) \mathcal{N}(\rho_K\vert \tilde s(T, z_K), I)$, $q(z_1, \rho_1) = q(z_1) \mathcal{N}(\rho_1\vert \tilde s(\delta, z_1))$, the forward transitions from \cref{alg:Fldvi} with exact momentum resampling for $\eta = 0$ (high friction limit), and the backward transition from \cref{alg:Bldvi} using the momentum resampling distribution described in \cref{app:proofs}.
\end{theorem}

\textbf{\UHA} \citep{UHA, dais} was developed as an approximation to Annealed Importance Sampling using underdamped Langevin dynamics. It uses unadjusted underdamped Langevin kernels for the forward transitions, and the unadjusted reversal of a Metropolis adjusted underdamped Langevin kernel for the backward transitions (simply put, Geffner and Domke \citep{UHA} and Zhang et al. \citep{dais} derived an exact expression for the reversal of a Metropolis adjusted underdamped Langevin kernel, and proposed to remove the correction step to define the backward transition).

\begin{theorem}\label{thm:uha}
\UHA\ is recovered by the formulation from \cref{sec:framework} with $\unnorm p(z_K, \rho_k) = \unnorm p(z_K) \mathcal{N}(\rho_K\vert 0, I)$, $q(z_1, \rho_1) = q(z_1) \mathcal{N}(\rho_1\vert 0, I)$, $s(t, z, \rho) = 0$, and the transitions from \cref{alg:Fldvi,,alg:Bldvi} with exact momentum resampling (possible due to removing the score network) with a learnable $\eta$.
\end{theorem}

We include proofs in \cref{app:proofs}. All follow similar steps, we get the exact transitions and expression for the ELBO given the specific choices made in each case, and compare to that of the original method, verifying their equivalence.

\subsection{LDVI: A New Method} \label{sec:newmethod}

Apart from recovering many existing methods, new algorithms can be derived using the proposed simulation scheme. As an example, we propose Langevin Diffusion VI (\LDVI), a novel method that combines the benefits of \MCD\ (backward transitions aided by a score network) with the benefits of \UHA\ (underdamped dynamics).
It is obtained by using the formulation from \cref{sec:framework} with $\unnorm p(z_K, \rho_K) = \unnorm p(z_K) \mathcal{N}(\rho_K\vert 0, I)$, $q(z_1, \rho_1) = q(z_1)\mathcal{N}(\rho_1 \vert 0, I)$, a full score network $s(t, z, \rho)$, and the transitions from \cref{alg:Fldvi,,alg:Bldvi} with the momentum resampling distributions
\begin{align*}
m_F(\rho_k' \vert \rho_k, \gamma, \delta) &= \mathcal{N}(\rho_k' \vert \rho_k(1 - \gamma \delta), 2\gamma \delta I)\\
m_B(\rho_k \vert \rho_k', z_k, \gamma, \delta) &= \!\begin{multlined}[t][4.5cm]
 \mathcal{N}(\rho_k \vert \rho_k'(1 - \gamma \delta) +  \\
  2\gamma\delta s(k\delta, z_k, \rho_k'), 2\gamma \delta I),
 \end{multlined}
\end{align*}
which are obtained by simulating components $\OS$ and $\OSB$ using the Euler-Maruyama scheme \citep{maruyama1955continuous, bayram2018numerical}.


\section{EXPERIMENTS} \label{sec:exps}

\begin{table*}[t]
\centering
\begin{tabular}{lcccccccc}
\toprule
 & \multicolumn{4}{c}{Ionosphere} & \multicolumn{4}{c}{Sonar} \\
\cmidrule(l{2pt}r{2pt}){2-5} \cmidrule(l{2pt}r{2pt}){6-9}
 & \ULA & \MCD & \UHA & \LDVI & \ULA & \MCD & \UHA & \LDVI \\
\midrule
$K=8$ & $-116.4\std{.05}$ & $-114.6\std{.01}$ & $-115.6\std{.05}$ & $\mathbf{-114.4}\std{.02}$ & $-122.4\std{.1}$ & $-117.2\std{.1}$ & $-120.1\std{.02}$ & $\mathbf{-116.3}\std{.03}$ \\
$K=16$ & $-115.4\std{.01}$ & $-113.6\std{.03}$ & $-114.4\std{.03}$ & $\mathbf{-113.1}\std{.01}$ & $-119.9\std{.02}$ & $-114.4\std{.02}$ & $-116.8\std{.08}$ & $\mathbf{-112.6}\std{.04}$ \\
$K=32$ & $-114.5\std{.05}$ & $-112.9\std{.05}$ & $-113.4\std{.03}$ & $\mathbf{-112.4}\std{.03}$ & $-117.4\std{.1}$ & $-112.4\std{.05}$ & $-113.9\std{.1}$ & $\mathbf{-110.6}\std{.08}$ \\
$K=64$ & $-113.8\std{.02}$ & $-112.5\std{.04}$ & $-112.8\std{.04}$ & $\mathbf{-112.1}\std{.01}$ & $-115.3\std{.05}$ & $-111.1\std{.7}$ & $-111.9\std{.02}$ & $\mathbf{-109.7}\std{.02}$ \\
$K=128$ & $-113.1\std{.04}$ & $-112.2\std{.02}$ & $-112.3\std{.02}$ & $\mathbf{-111.9}\std{.01}$ & $-113.5\std{.03}$ & $-110.2\std{.04}$ & $-110.6\std{.07}$ & $\mathbf{-109.1}\std{.03}$ \\
$K=256$ & $-112.7\std{.01}$ & $-112.1\std{.02}$ & $-112.1\std{.02}$ & $\mathbf{-111.7}\std{.01}$ & $-112.1\std{.08}$ & $-109.7\std{.03}$ & $-109.7\std{.05}$ & $\mathbf{-108.9}\std{.02}$ \\
\bottomrule
\end{tabular}
\caption{\textbf{Combining underdamped dynamics with score networks, as done by \LDVI, yields better results than all other methods for both datasets.} ELBO (higher is better, standard deviations in \textcolor{\cstd}{\cstd}) achieved after training by different methods for different values of $K$ for a logistic regression model with two datasets, \textit{ionosphere} ($d=35$) and \textit{sonar} ($d=61$). Plain VI achieves an ELBO of $-124.1\std{.15}$ nats with the \textit{ionosphere} dataset, and $-138.6\std{.2}$ nats with the \textit{sonar} dataset. Best result for each dataset and value of $K$ highlighted.}
\label{tab:lr}
\end{table*}

\begin{table*}[]
\centering
\begin{tabular}{lcccccccc}
\toprule
 & \multicolumn{4}{c}{Brownian motion} & \multicolumn{4}{c}{Lorenz system} \\
\cmidrule(l{2pt}r{2pt}){2-5} \cmidrule(l{2pt}r{2pt}){6-9}
 & \ULA & \MCD & \UHA & \LDVI & \ULA & \MCD & \UHA & \LDVI \\
\midrule
$K=8$ & $-1.9\std{.05}$ & $-1.4\std{.06}$ & $-1.6\std{.03}$ & $\mathbf{-1.1}\std{.03}$ & $-1168.2\std{.1}$ & $-1168.1\std{.1}$ & $-1166.3\std{.1}$ & $\mathbf{-1166.1}\std{.06}$ \\
$K=16$ & $-1.5\std{.06}$ & $-0.8\std{.04}$ & $-1.1\std{.03}$ & $\mathbf{-0.5}\std{.03}$ & $-1165.7\std{.1}$ & $-1165.6\std{.1}$ & $-1163.1\std{.3}$ & $\mathbf{-1162.2}\std{.07}$ \\
$K=32$ & $-1.1\std{.05}$ & $-0.4\std{.05}$ & $-0.5\std{.04}$ & $\mathbf{0.1}\std{.04}$ & $-1163.2\std{.04}$ & $-1163.3\std{.04}$ & $-1160.3\std{.05}$ & $\mathbf{-1157.6}\std{.1}$ \\
$K=64$ & $-0.7\std{.03}$ & $-0.1\std{.1}$ & $0.1\std{.02}$ & $\mathbf{0.5}\std{.03}$ & $-1160.9\std{.04}$ & $-1161.1\std{.04}$ & $-1157.7\std{.05}$ & $\mathbf{-1153.7}\std{.1}$ \\
$K=128$ & $-0.3\std{.03}$ & $0.2\std{.04}$ & $0.4\std{.01}$ & $\mathbf{0.7}\std{.01}$ & $-1158.9\std{.05}$ & $-1158.9\std{.05}$ & $-1155.4\std{.07}$ & $\mathbf{-1153.1}\std{.1}$ \\
$K=256$ & $-0.1\std{.02}$ & $0.5\std{.01}$ & $0.6\std{.01}$ & $\mathbf{0.9}\std{.02}$ & $-1157.2\std{.06}$ & $-1157.1\std{.06}$ & $-1153.3\std{.1}$ & $\mathbf{-1151.1}\std{.2}$ \\
\bottomrule
\end{tabular}
\caption{\textbf{\LDVI\ yields better results than all other methods for both models.} ELBO (higher is better, standard deviations in \textcolor{\cstd}{\cstd}) achieved after training by different methods for different values of $K$ for two time series models ($d=32$ for the Brownian motion model and $d=90$ for Lorenz system). Plain VI achieves an ELBO of $-4.4\std{.02}$ nats on the Brownian motion model and $-1187.8\std{.4}$ nats on the Lorenz system model. Best result for each model and value of $K$ highlighted.}
\label{tab:ts}
\end{table*}

\begin{table}[h]
\centering
\begin{tabular}{lcccc}
\toprule
& \multicolumn{4}{c}{Random effect regression (seeds)} \\
\cmidrule(l{2pt}r{2pt}){2-5}
 & \ULA & \MCD & \UHA & \LDVI \\
\midrule
$K=8$& $-75.5\std{.02}$ & $-75.1\std{.05}$ & $\mathbf{-74.9}\std{.01}$ & $\mathbf{-74.9}\std{.01}$ \\
$K=16$& $-75.2\std{.04}$ & $-74.6\std{.04}$ & $-74.6\std{.01}$ & $\mathbf{-74.5}\std{.03}$ \\
$K=32$& $-74.9\std{.03}$ & $-74.3\std{.03}$ & $\mathbf{-74.2}\std{.02}$ & $\mathbf{-74.2}\std{.02}$ \\
$K=64$& $-74.6\std{.01}$ & $-74.1\std{.02}$ & $-74.1\std{.01}$ & $\mathbf{-73.9}\std{.05}$ \\
$K=128$& $-74.3\std{.03}$ & $-73.9\std{.01}$ & $-73.8\std{.02}$ & $\mathbf{-73.7}\std{.01}$ \\
$K=256$& $-74.1\std{.01}$ & $-73.7\std{.01}$ & $-73.7\std{.02}$ & $\mathbf{-73.6}\std{.01}$ \\
\bottomrule
\end{tabular}
\caption{\textbf{\LDVI\ yields better results than all other methods.} ELBO (higher is better, standard deviations in \textcolor{\cstd}{\cstd}) achieved after training by different methods for different values of $K$ for a random effect regression model with the \textit{seeds} dataset ($d=26$). Plain VI achieves an ELBO of $-77.1\std{.02}$ nats. Best result for each value of $K$ highlighted.}
\label{tab:sds}
\end{table}

\begin{table*}[h]
\centering
\begin{tabular}{lcccccccc}
\toprule
 & \multicolumn{4}{c}{Brownian motion} & \multicolumn{4}{c}{Logistic regression (sonar)} \\
\cmidrule(l{2pt}r{2pt}){2-5} \cmidrule(l{2pt}r{2pt}){6-9}
 & \UHAEM & \UHA & \LDVIEM & \LDVI & \UHAEM & \UHA & \LDVIEM & \LDVI \\
\midrule
$K=8$ & $-2.8\std{.4}$ & $-1.6\std{.03}$ & $-2.8\std{.4}$ & $\mathbf{-1.1}\std{.03}$ & $-124.1\std{.1}$ & $-120.1\std{.02}$ & $-118.5\std{.1}$ & $\mathbf{-116.3}\std{.03}$ \\
$K=16$ & $-2.2\std{.04}$ & $-1.1\std{.03}$ & $-1.4\std{.03}$ & $\mathbf{-0.5}\std{.03}$ & $-119.9\std{.08}$ & $-116.8\std{.08}$ & $-114.4\std{.05}$ & $\mathbf{-112.6}\std{.04}$ \\
$K=32$ & $-1.6\std{.02}$ & $-0.5\std{.04}$ & $-0.5\std{.02}$ & $\mathbf{0.1}\std{.04}$ & $-116.4\std{.1}$ & $-113.9\std{.1}$ & $-111.7\std{.04}$ & $\mathbf{-110.6}\std{.08}$ \\
$K=64$ & $-0.9\std{.04}$ & $0.1\std{.02}$ & $0.1\std{.05}$ & $\mathbf{0.5}\std{.03}$ & $-113.8\std{.1}$ & $-111.9\std{.02}$ & $-110.3\std{.04}$ & $\mathbf{-109.7}\std{.02}$ \\
$K=128$ & $-0.4\std{.03}$ & $0.4\std{.01}$ & $0.4\std{.04}$ & $\mathbf{0.7}\std{.01}$ & $-111.9\std{.1}$ & $-110.6\std{.07}$ & $-109.6\std{.04}$ & $\mathbf{-109.1}\std{.03}$ \\
$K=256$ & $0.1\std{.04}$ & $0.6\std{.01}$ & $0.6\std{.05}$ & $\mathbf{0.9}\std{.02}$ & $-110.7\std{.1}$ & $-109.7\std{.05}$ & $-109.1\std{.06}$ & $\mathbf{-108.9}\std{.02}$ \\
\bottomrule
\end{tabular}
\caption{ELBO (higher is better, standard deviations in \textcolor{\cstd}{\cstd}) achieved by \UHAEM\ and \LDVIEM, the variants of \UHA\ and \LDVI\ that use the transitions from \cref{alg:fsimple,,alg:bwsimple}.}
\label{tab:direct}
\end{table*}

This section presents an empirical evaluation of different methods that follow our framework. We are interested in the effect that different choices have on the final performance. Specifically, we are interested in studying the benefits of using underdamped dynamics (\UHA\ and \LDVI) instead of the overdamped variant (\ULA\ and \MCD),
the benefits of using powerful backward transitions aided by learnable score networks (\MCD\ and \LDVI),
and the benefits of combining both improvements (\LDVI) against each individual one (\MCD\ and \UHA).
We explore this empirically in \cref{sec:exps1}.

We are also interested in how the numerical simulation scheme used affects the methods' performance. We explore this in \cref{sec:naive}, where we propose and evaluate empirically an alternative simulation scheme based on a simpler splitting than the one introduced in \cref{sec:discr}.

In all cases, we use the different methods to perform inference on a wide range of tasks for values of $K \in \{8, 16, 32, 64, 128, 256\}$, and report mean ELBO achieved after training, with standard deviations computed over three different random seeds. For all methods we set $q(z)$ to a mean-field Gaussian, initialized to a maximizer of the ELBO, and train all parameters using Adam for 150000 steps. We repeat all simulations for the learning rates $10^{-3}$, $10^{-4}$ and $10^{-5}$, and keep the best one for each method and model. For all methods we tune the initial distribution $q(z)$, discretization step-size $\delta$, and the bridging densities' parameters $\beta$. For \LDVI\ and \UHA\ we also tune the damping coefficient $\gamma > 0$, and for \LDVI\ and \MCD\ we tune the score network $s$, which has two hidden layers with residual connections \citep{he2016deep}. We implement all methods using Jax \citep{jax2018github}.

\subsection{Underdamped dynamics and score networks} \label{sec:exps1}

\paragraph{Logistic Regression.} \Cref{tab:lr} shows results achieved by \ULA, \MCD, \UHA\ and \LDVI\ on a logistic regression model with two datasets, \textit{ionosphere} \citep{sigillito1989classification} ($d=35$) and \textit{sonar} \citep{gorman1988analysis} ($d=61$). It can be observed that going from overdamped to underdamped dynamics yields significant performance improvements: For the \textit{sonar} dataset, \UHA\ with $K=64$ bridging densities performs better than \ULA\ with $K=256$, and \LDVI\ with $K=64$ performs better than \MCD\ with $K=256$. Similarly, it can be observed that the use of score networks for the backward transitions also yields significant gains: For the \textit{sonar} dataset, \MCD\ and \LDVI\ with $K=64$ outperform \ULA\ and \UHA, respectively, with $K=256$. Finally, results show that combining both improvements is beneficial as well, as it can be seen that \LDVI\ outperforms all other methods for all datasets and values of $K$.

\paragraph{Time series.} We consider two time series models obtained by discretizing two different SDEs, one modeling a Brownian motion with a Gaussian observation model ($d=32$), and other modeling a Lorenz system, a three-dimensional nonlinear dynamical system used to model atmospheric convection ($d=90$). We give details for these models in \cref{app:models}. Both were obtained from the ``Inference Gym'' \citep{inferencegym2020}. Results are shown in \cref{tab:ts}. The conclusions are similar to the ones for the logistic regression models: underdamped dynamics and score networks yield gains in performance, and \LDVI, which combines both improvements, performs better than all other methods.

\paragraph{Random effect regression.} \Cref{tab:sds} shows results on a random effect regression model with the \textit{seeds} dataset \citep{crowder1978beta} ($d=26$) (model details in \cref{app:models}). The same conclusions hold, both \UHA\ and \MCD\ perform better than \ULA, and \LDVI\ performs better than all other methods.

\subsection{Effect of Numerical Simulation Scheme} \label{sec:naive}

All the methods studied so far are based on the simulation scheme introduced in \cref{sec:discr}. We note that other simulations methods, which yield different transitions, could be used. We are interested in studying how the simulation scheme used affects methods' performance.

We consider the forward and backward transitions shown in \cref{alg:fsimple,,alg:bwsimple}, obtained by simulating the SDEs using the Euler-Maruyama scheme as explained in \cref{app:dersimpler} (where we also give an expression for the ratio of the transitions). We propose two methods using these transitions: One that uses a full score network $s$, and other one that uses no score network (i.e. fixes $s=0$). Intuitively, these can be seen as variants of \LDVI\ and \UHA\ that use this new simulation scheme, so we term them \LDVIEM\ and \UHAEM.\footnote{Using this simulation it is not directly clear how to take the high-friction limit. Thus, it is unclear whether they can be used to recover methods analogous to \ULA\ and \MCD.}

\begin{algorithm}[ht]
\caption{Forward transition $F_k(z_{k+1}, \rho_{k+1} \vert z_{k}, \rho_{k})$ obtained with modified Euler-Maruyama}
\label{alg:fsimple}
\begin{algorithmic}
\State Resample momentum $\rho_{k+1} \sim \mathcal{N}(\rho_k (1 - \gamma \delta) + \delta \nabla \log \pi_{k\delta}(z_k), 2\gamma\delta I)$.
\State Update position $z_{k+1} = z_k + \delta \rho_{k+1}$.
\State \Return $(z_{k+1}, \rho_{k+1})$
\end{algorithmic}
\end{algorithm}

\begin{algorithm}[ht]
\caption{Backward transition $B_k(z_{k}, \rho_{k} \vert z_{k+1}, \rho_{k+1})$ obtained with modified Euler-Maruyama}
\label{alg:bwsimple}
\begin{algorithmic}
\State Update position $z_k = z_{k+1} - \delta \rho_{k+1}$.
\State Resample momentum $\rho_k \sim \mathcal{N}(\rho_{k+1}(1-\delta\gamma) - \delta \nabla \log \pi_{k\delta}(z_{k}) + 2\delta \gamma s(k\delta, z_k, \rho_{k+1}), 2\delta\gamma I)$.
\State \Return $(z_{k}, \rho_{k})$
\end{algorithmic}
\end{algorithm}

\Cref{tab:direct} shows results for the Brownian motion model and the logistic regression model with the \textit{sonar} dataset (full results in \cref{app:dersimpler}). It can be observed that each of \LDVIEM\ and \UHAEM\ performs worse than its counterpart using the simulation scheme from \cref{sec:discr}, \LDVI\ and \UHA. Interestingly, for the Brownian motion model, \UHAEM\ is also outperformed by \ULA. This sheds light on the importance of the simulation scheme used: for some models, the benefits obtained by using underdamped dynamics and a score network may be lost by using a poorly chosen simulation scheme.


\bibliography{ref}
\bibliographystyle{plainnat}

\clearpage
\newpage
\appendix

\onecolumn

\section{MODEL DETAILS} \label{app:models}

\subsection{Time Series Models}

This models are presented following their descriptions in the Inference Gym documentation \citep{inferencegym2020}. Both are obtained by discretizing an SDE and using a Gaussian observation model.

\paragraph{Brownian motion with Gaussian observation noise} The model is given by
\begin{align*}
\alpha_\mathrm{inn} & \sim \mathrm{LogNormal}(\mathrm{loc} = 0, \mathrm{scale} = 2)\\
\alpha_\mathrm{obs} & \sim \mathrm{LogNormal}(\mathrm{loc} = 0, \mathrm{scale} = 2)\\
x_1 & \sim \mathcal{N}(\mathrm{loc} = 0, \mathrm{scale} = \alpha_\mathrm{inn})\\
x_i & \sim \mathcal{N}(\mathrm{loc} = x_{i-1}, \mathrm{scale} = \alpha_\mathrm{inn}) & i=2,\hdots,30\\
y_i & \sim \mathcal{N}(\mathrm{loc} = x_{i}, \mathrm{scale} = \alpha_\mathrm{obs}) & i=1,\hdots,30.
\end{align*}
The goal is to do inference over variables $\alpha_\mathrm{inn}$, $\alpha_\mathrm{obs}$ and $x_i$ ($i=1,\hdots,30$), given the observations $y_i$, for $i \in \{1, \hdots, 10\} \cup \{20, \hdots, 30\}$ (i.e. the ten middle observations are missing).

\paragraph{Lorenz system} The model is given by
\begin{align*}
x_1 & \sim \mathcal{N}(\mathrm{loc} = 0, \mathrm{scale} = 1)\\
y_1 & \sim \mathcal{N}(\mathrm{loc} = 0, \mathrm{scale} = 1)\\
z_1 & \sim \mathcal{N}(\mathrm{loc} = 0, \mathrm{scale} = 1)\\
x_i & \sim \mathcal{N}(\mathrm{loc} = 10(y_{i-1} - x_{i-1}), \mathrm{scale} = \alpha_{\mathrm{inn}}) & i=2,\hdots,30\\
y_i & \sim \mathcal{N}(\mathrm{loc} = x_{i-1}\, (28 - z_{i-1}) - y_{i-1}), \mathrm{scale} = \alpha_{\mathrm{inn}}) & i=2,\hdots,30\\
z_i & \sim \mathcal{N}(\mathrm{loc} = x_{i-1}\, y_{i-1} - \frac{8}{3} \, z_{i-1} , \mathrm{scale} = \alpha_{\mathrm{inn}}) & i=2,\hdots,30,\\
o_i & \sim \mathcal{N}(\mathrm{loc} = x_{i}, \mathrm{scale} = 1) & i=2,\hdots,30,
\end{align*}
where $\alpha_{\mathrm{inn}} = 0.1$ (determined by the discretization step-size used for the original SDE). The goal is to do inference over $x_i, y_i, z_i$ for $i=1,\hdots,30$, given observed values $o_i$ for $i \in \{1, \hdots, 10\} \cup \{20, \hdots, 30\}$.

\subsection{Random effect regression} This model can be found in the MultiBUGS \citep{goudie2020multibugs} documentation. It is essentially a random effects regression model, given by
\begin{align*}
\tau & \sim \mathrm{Gamma}(0.01, 0.01)\\
a_0 & \sim \mathcal{N}(0, 10)\\
a_1 & \sim \mathcal{N}(0, 10)\\
a_2 & \sim \mathcal{N}(0, 10)\\
a_{12} & \sim \mathcal{N}(0, 10)\\
b_i & \sim \mathcal{N}\left(0, \frac{1}{\sqrt{\tau}}\right) & i=1,\hdots, 21\\
\mathrm{logits}_i & = a_0 + a_1 \, x_i + a_2 \, y_i + a_{12} \, x_i \, y_i + b_1 & i=1,\hdots, 21\\
r_i & \sim \mathrm{Binomial}(\mathrm{logits}_i, N_i) & i=1,\hdots, 21.
\end{align*}
The goal is to do inference over the variables $\tau, a_0, a_1, a_2, a_{12}$ and $b_i$ for $i=1,\hdots,21$, given observed values for $x_i$, $y_i$ and $N_i$. The data used was obtained from \citep{crowder1978beta}, which models the germination proportion of seeds arranged in a factorial layout by seed and type of root.

\newpage
\section{TIME-REVERSED SDE FOR UNDERDAMPED LANGEVIN PROCESS} \label{app:reverseSDE}

Consider a diffusion process of the form
\begin{equation}
\dd{x^t} = f(x^t, t) \dd t + G(t) \dd{w^t},
\end{equation}
where $t\in {0, T}$. Defining $u^t = x^{T-t}$, the process that inverts the one above is given by \citep{haussmann1986time, anderson1982reverse, dockhorn2021score}
\begin{equation} \label{eq:revsdegen}
\dd{u^t} = \left[ -f(u^t, T-t) + G(T-t) G(T-t)^\top \nabla \log q^{T-t}(u^t)\right] \dd t + G(T-t) \dd{w^t}.
\end{equation}

This result can be used to derive the time-reversed diffusion for the underdamped Langevin process, by expressing the forward process as (using $x = [z, \rho]$)
\begin{equation}
    \left[\begin{array}{l}
    \dd{z^t}\\
    \dd{\rho^t}
    \end{array}\right] = 
    \underbrace{\left[\begin{array}{c}
    \rho^t\\
    \nabla \log \pi^t(z^t) - \gamma \rho^t
    \end{array}\right]}_{f(x^t, t)} \dd{t} + 
    \underbrace{\left[\begin{array}{cc}
    0 & 0\\
    0 & \sqrt{2\gamma}
    \end{array}\right]}_{G(t)}
    \left[\begin{array}{l}
    \dd{w_1^t}\\
    \dd{w_2^t}
    \end{array}\right],
\end{equation}
and applying \cref{eq:revsdegen}.

\section{DERIVATION OF BACWARD TRANSITIONS FROM ALGORITHM \ref{alg:Bldvi}} \label{app:derivingB}

The time-reversed SDE is split into three components as
\begin{equation*}
\left[\begin{array}{c}
\dd y^t\\
\dd \lambda^t
\end{array}\right] = 
\underbrace{\left[\begin{array}{c}
-\lambda^t \dd t\\
0
\end{array}\right]}_{\ASB} + 
\underbrace{\left[\begin{array}{c}
0\\
-\nabla \log \pi^{T-t}(y^t) \dd t
\end{array}\right]}_{\BSB} + 
\underbrace{\left[\begin{array}{c}
0\\
-\gamma \rho^t \dd t + 2\gamma s(T-t, y^t, \lambda^t) \dd t + \sqrt{2\gamma} \dd w^t
\end{array}\right]}_{\OSB}.
\end{equation*}
Then, the final transitions are given by sequentially composing the simulations for components $\BSB\ASB\BSB\OSB$.
\begin{description}
    \item[Simulating $\ASB$:] This can be done exactly. Given initial values $(y^{t_0}, \lambda^{t_0})$ at time $t_0$, simulating $\ASB$ for a time $\delta$ results in $(y^{t_0+\delta}, \lambda^{t_0+\delta}) = (y^{t_0} + \delta \lambda^{t_0}, \lambda^{t_0})$.
    \item[Simulating $\BSB$:] Given initial values $(y^{t_0}, \lambda^{t_0})$ at time $t_0$, and using that $\pi^{T-t_0} \approx \pi^{T-t_0+\delta}$ for small $\delta$, simulating $\BS$ for a time $\delta$ results in $(y^{t_0+\delta}, \lambda^{t_0+\delta}) = (y^{t_0}, \lambda^{t_0} + \delta \nabla \log \pi^{T-t_0}(y^{t_0}))$.
    \item[Simulating $\OSB$:] In contrast to the forward transitions, this component cannot be simulated exactly due to the presence of the term involving $s$ (it can be done exactly if we fix $s=0$). This component can be simulated approximately, for instance, using the Euler-Maruyama scheme. Given an initial values $y^{t_0}$ and $\lambda^{t_0}$ at time $t_0$, simulating $\OSB$ for a time $\delta$ gives $\lambda^{t_0+\delta} \sim \mathcal{N}(\lambda^{t_0+\delta} \vert \lambda^{t_0}(1 - \gamma\delta) + 2\gamma\delta s(T-t, y^{t_0}), 2\gamma\delta I)$. We will use $m_B$ to denote generically the momentum resampling distribution used.
\end{description}
Finally, the backward transitions from \cref{alg:Bldvi} are obtained by sequentially combining the results above, and transforming back to the variables $z$ and $\rho$ using the fact that $z^t = y^{T-t}$ and $\rho^t = \lambda^{T-t}$.

\newpage

\section{PROOFS} \label{app:proofs}

\subsection{Proof of \cref{lemma:ratio}}

This proof follows closely the one from Geffner and Domke \citep{UHA}. We will derive expressions for the forward and backward transitions separately, and take the ratio at the end. In the derivation, we replace all delta functions by Gaussians with variance $\Delta$, and take the limit $\Delta \to 0$ with the final expressions. We use $(z', \rho') = \LPk(z, \rho)$ to denote the leapfrog integration step typically used by HMC targeting the bridging density $\pi_k$ (see \cref{alg:Fldvi} for a definition), which is invertible and volume preserving (Jacobian determinant equals one).

\paragraph{Forward transitions} We will divide the forward transition from \cref{alg:Fldvi} in two steps:
\begin{enumerate}
    \item Resampling step: $z_k', \rho_k' \sim m_F(\rho_k'\vert \rho_k, \gamma, \delta) \, \mathcal{N}(z_k' \vert z_k, \Delta)$,
    \item Deterministic step: $(z_{k+1}, \rho_{k+1}) = \LPk(z_k', \rho_k')$.
\end{enumerate}
Using that $\LPk$ is invertible and volume preserving, we get
\begin{equation}
F_k(z_{k+1}, \rho_{k+1}\vert z_k, \rho_k) = m_F\left((\LPk^{-1})^\rho(z_{k+1}, \rho_{k+1}) \Big\vert \rho_k, \gamma, \delta\right) \mathcal{N}\left((\LPk^{-1})^z(z_{k+1}, \rho_{k+1}) \Big\vert z_k, \Delta\right),
\end{equation}
where $(\LPk^{-1})^\rho(z_{k+1}, \rho_{k+1})$ is defined as the second component of $\LPk^{-1}(z_{k+1}, \rho_{k+1})$, and similarly $(\LPk^{-1})^z(z_{k+1}, \rho_{k+1})$ as its first component.

\paragraph{Backward transitions} Similarly, we divide the backward transitions from \cref{alg:Bldvi} in two steps:
\begin{enumerate}
    \item Deterministic step: $(z_{k}', \rho_{k}') = \LPk^{-1}(z_{k+1}, \rho_{k+1})$,
    \item Resampling step: $z_k, \rho_k \sim \mathcal{N}(z_k \vert z_k', \Delta) \, m_B(\rho_k\vert \rho_k', z_k, \gamma, \delta)$.
\end{enumerate}
Then, we get
\begin{equation}
B_k(z_k, \rho_k \vert z_{k+1}, \rho_{k+1}) = \mathcal{N}\left(z_k \Big\vert (\LPk^{-1})^z(z_{k+1}, \rho_{k+1}), \Delta\right) \, m_B\left(\rho_k\Big\vert (\LPk^{-1})^\rho(z_{k+1}, \rho_{k+1}), z_k, \gamma, \delta\right).
\end{equation}

Finally, using $\rho_k' = (\LPk^{-1})^\rho(z_{k+1}, \rho_{k+1})$ and $z_k' = (\LPk^{-1})^z(z_{k+1}, \rho_{k+1})$ to simplify notation, taking the ratio between $F_k$ and $B_k$ yields
\begin{equation}
\frac{B_k(z_k, \rho_k \vert z_{k+1}, \rho_{k+1})}{F_k(z_{k+1}, \rho_{k+1}\vert z_k, \rho_k)} = \frac{m_B\left(\rho_k\vert \rho_k', z_k, \gamma, \delta\right)}{m_F(\rho_k'\vert \rho_k, \gamma, \delta)},
\end{equation}
since the ratio between the Gaussian densities cancel.

\newpage

\subsection{Proof of \cref{thm:ula}}

\paragraph{\ULA\ summary} \ULA\ uses the following transitions
\begin{equation}
\begin{split}
F_k(z_{k+1} \vert z_k) &= \mathcal{N}(z_{k+1} \vert z_k + \epsilon \nabla \log \pi_{k}(z_k), 2\epsilon)\\
B_k(z_k \vert z_{k+1}) &= \mathcal{N}(z_k \vert z_{k+1} + \epsilon \nabla \log \pi_{k}(z_{k+1}), 2\epsilon)
\end{split}
\end{equation}

The forward transition yields 
\begin{equation} \label{eq:ULAforward}
z_{k+1} = z_k + \epsilon \nabla \log \pi_{k}(z_k) + \sqrt{2 \epsilon} \, \xi_k,
\end{equation}
where $\xi_k \sim \mathcal{N}(0, I)$, for $k=1, ..., K-1$.

Then, ULA's lower bound can be written as (using $\mathcal{N}(a\vert b, c)$ with $c \geq 0$ to denote the pdf of a Gaussian with mean $b$ and variance $c$ evaluated at $a$)
\begin{align}
\frac{p(z_K)}{q(z_1)} \prod_{k=1}^{K-1} \frac{B_k(z_k\vert z_{k+1})}{F_k(z_{k+1} \vert z_k)} & = \frac{p(z_K)}{q(z_1)} \prod_{k=1}^{K-1} \frac{\mathcal{N}(z_k \vert z_{k+1} + \epsilon \nabla \log \pi_{k}(z_{k+1}), 2\epsilon)}{\mathcal{N}(z_{k+1} \vert z_k + \epsilon \nabla \log \pi_{k}(z_k), 2\epsilon)}\\
& = \frac{p(z_K)}{q(z_1)} \prod_{k=1}^{K-1} \frac{\mathcal{N}(- \epsilon \nabla \log \pi_{k}(z_k) - \epsilon \nabla \log \pi_{k}(z_{k+1}) - \sqrt{2\epsilon} \, \xi_k \vert 0, 2\epsilon)}{\mathcal{N}(\sqrt{2\epsilon} \, \xi_k\vert 0, 2\epsilon)}\label{eq:ULAREF}\\
& = \frac{p(z_K)}{q(z_1)} \prod_{k=1}^{K-1} \frac{\mathcal{N}\left( \sqrt{\frac{\epsilon}{2}} \log \pi_{k}(z_k) + \sqrt{\frac{\epsilon}{2}} \nabla \log \pi_{k}(z_{k+1}) + \xi_k \vert 0, 1\right)}{\mathcal{N}(\xi_k\vert 0, 1)}, \label{eq:ULAlowerbound}
\end{align}
where we obtain \cref{eq:ULAREF} using the expression for $z_{k+1}$ from \cref{eq:ULAforward}.

\paragraph{Recovering \ULA} Using exact momentum resampling for the forward transitions gives $\rho_{k}' \sim \mathcal{N}(\eta \rho_k, (1 - \eta^2) I)$, where $\eta = \exp(-\gamma \delta)$.) Using $\eta = 0$ (high friction limit) gives $\rho_k' \sim \mathcal{N}(0, I)$, which is used in place of $m_F(\rho_k'\vert \rho_k, \gamma, \delta)$ in \cref{alg:Fldvi}. The final forward transitions $F_k(z_{k+1}, \rho_{k+1} \vert z_{k}, \rho_{k})$ are thus given by
\begin{equation}
\begin{alignedat}{2} \label{eq:UHAforwardULA}
\mbox{Simulate $\OS$:}\qquad \rho_k' & \sim \mathcal{N}(0, I) & & \qquad \rightarrow \rho_k'=\xi_k \sim \mathcal{N}(0, I)\\
\mbox{Simulate $\BS$:}\qquad \rho_k'' & = \rho_k' + \frac{\delta}{2} \nabla \log \pi_{k}(z_k) & & \qquad \rightarrow  \rho_k'' = \xi_k + \frac{\delta}{2} \nabla \log \pi_{k}(z_k)\\
\mbox{Simulate $\AS$:}\quad z_{k+1} & = z_k + \delta \rho_k'' & & \qquad \rightarrow z_{k+1} = z_k + \frac{\delta^2}{2} \nabla \log \pi_k(z_k) + \delta \xi_k\\
\mbox{Simulate $\BS$:}\quad \rho_{k+1} & = \rho_k'' + \frac{\delta}{2} \nabla \log \pi_k(z_{k+1}) & &\qquad \rightarrow \rho_{k+1} = \xi_k + \frac{\delta}{2} \nabla \log \pi_k(z_k) + \frac{\delta}{2} \nabla \log \pi_k(z_{k+1}).
\end{alignedat}
\end{equation}
It can be seen that the forward dynamics for $z$ are exactly the same as those used by \ULA\ (\cref{eq:ULAforward}), using $\epsilon = \nicefrac{\delta^2}{2}$.

Thanks to removing the score network (i.e. $s(t, z, \rho) = 0$), exact momentum resampling is possible for the backward transitions $B_k(z_{k}, \rho_{k} \vert z_{k+1}, \rho_{k+1})$ as well. Similarly to the forward transitions, this gives $\rho_k \sim \mathcal{N}(0, I)$, which should be used in place of $m_B(\rho_k\vert \rho_k', z_k, \gamma, \delta)$ in \cref{alg:Bldvi}.

Finally, using the transitions with these resampling distributions, the momentum augmented distributions given by $\unnorm p(z_K, \rho_k) = \unnorm p(z_K) \mathcal{N}(\rho_K\vert 0, I)$ and $q(z_1, \rho_1) = q(z_1) \mathcal{N}(\rho_1\vert 0, I)$, and the result from \cref{lemma:ratio}, we get
\begin{align}
\frac{p(\zg, \rhog)}{q(\zg, \rhog)} & = \frac{p(z_K)}{q(z_1)} \prod_{k=1}^{K-1} \frac{\mathcal{N}(\rho_{k+1} \vert 0, I)}{\mathcal{N}(\rho_k'\vert 0, I)}\\
& = \frac{p(z_K)}{q(z_1)} \prod_{k=1}^{K-1} \frac{\mathcal{N}(\xi_k + \frac{\delta}{2} \nabla \log \pi_k(z_k) + \frac{\delta}{2} \nabla \log \pi_k(z_{k+1}) \vert 0, I)}{\mathcal{N}(\xi_k \vert 0, I)},
\end{align}
where the second line is obtained by replacing $\rho_{k+1}$ and $\rho_k'$ by their respective expressions from \cref{eq:UHAforwardULA}. Taking $\epsilon = \nicefrac{\delta^2}{2}$ gives the ratio used by \ULA\ (\cref{eq:ULAlowerbound}), showing that our framework with the choices from \cref{thm:ula} recovers \ULA.

\newpage

\subsection{Proof of \cref{thm:mcd}}

\paragraph{\MCD\ summary} \MCD\ uses the same forward transitions as ULA
\begin{equation}
F_k(z_{k+1}\vert z_k) = \mathcal{N}(z_k + \epsilon \nabla \log \pi_{k}(z_k), 2\epsilon),
\end{equation}
and backward transitions given by 
\begin{equation}
B_k(z_k \vert z_{k+1}) = \mathcal{N}\left(z_{k+1} + \epsilon \nabla \log \pi_{k}(z_{k+1}) + 2\epsilon \tilde s\left(k\delta, z_{k+1} \right), 2\epsilon\right).
\end{equation}

Using these transitions, and writing $z_{k+1} = z_k + \epsilon \nabla \log \pi_{k}(z_k) + \sqrt{2 \epsilon} \, \xi_k$ (where $\xi_k\sim \mathcal{N}(0, I)$), \MCD\ yields the ratio
\begin{multline}
\frac{\unnorm p(\zg)}{q(\zg)} = \frac{p(z_K)}{q(z_1)} \prod_{k=1}^{K-1} \frac{B_k(z_k\vert z_{k+1})}{F_k(z_{k+1} \vert z_k)}\\
=\frac{\unnorm p(z_K)}{q(z_1)} \prod_{k=1}^{K-1} \frac{\mathcal{N}(\sqrt{\frac{\epsilon}{2}} \nabla \log \pi_{k}(z_{k}) + \sqrt{\frac{\epsilon}{2}} \nabla \log \pi_{k}(z_{k+1}) + \sqrt{2\epsilon} \tilde s(k\delta, z_{k+1}) + \xi_k\vert 0, 1)}{\mathcal{N}(\xi_k \vert 0, 1)}.
\end{multline}

\paragraph{Recovering \MCD}

The forward transitions used by our framework to recover \MCD\ are exactly the same as the ones used in the proof of \cref{thm:ula}, shown in \cref{eq:UHAforwardULA}. Therefore, the forward dynamics for $z$ are given by $z_{k+1} = z_k + \frac{\delta^2}{2} \nabla \log \pi_{k}(z_k) + \delta \xi_k$, which is exactly the same as the forward dynamics used by \MCD, taking $\epsilon = \nicefrac{\delta^2}{2}$.

Deriving the backward transitions requires simulating component $\OSB$. Using $s(T-t, y_t, \lambda_t) = s(T-t, y_t)$ (as stated in \cref{thm:mcd}), this component is given by
\begin{equation}
\left[\begin{array}{c}
\dd y^t\\
\dd \lambda^t
\end{array}\right] = 
\left[\begin{array}{c}
0\\
-\gamma \rho^t \dd t + 2\gamma s(T-t, y^t) \dd t + \sqrt{2\gamma} \dd w^t
\end{array}\right].
\end{equation}
Since $\dd y^t = 0$, we get that $y^t$ is constant as a function of $t$. However, the term $s(T-t, y^t)$ is not a constant with respect to time, due to its first argument. Thus, in general, exact simulation for this component is intractable. However, approximating $s(T-t, y^t) \approx s(T-t_0, y^{t_0})$ for $t\in [t_0, t_0+\delta]$, we can simulate it as $\lambda^{t_0+\delta} \sim \mathcal{N}(\eta \lambda^{t_0} + 2s(T-t_0, y^t) (1 - \eta), (1 - \eta^2) I)$, where $\eta = \exp(-\gamma \delta)$.\footnote{This is obtained by noting that the process $\dd \lambda^t = a \dd t - \gamma \lambda^t \dd t + \sqrt{2\gamma} \dd w_t$, where $a$ is a constant, admits exact simulation as $\lambda^{t_0+\delta} \sim \mathcal{N}(\lambda^{t_0} \eta + \frac{a}{\gamma}(1 - \eta), (1-\eta^2)I)$, where $\eta = \exp(-\gamma \delta)$. Our result follows from setting $a = \gamma s(T-t_0, y^{t_0})$.} Taking $\eta=0$ and expressing the transitions in terms of $z$ and $\rho$ yields the backward transitions $B_k(z_{k}, \rho_{k} \vert z_{k+1}, \rho_{k+1})$ from \cref{alg:Bldvi} with momentum resampling distribution given by $m_B(\rho_k\vert \rho_k', z_k, \gamma, \delta) = \mathcal{N}\left(2 s(k\delta, z_k), I\right)$.

Using the above transitions, the momentum augmented distributions given by
\begin{equation*}
\unnorm p(z_K, \rho_K) = \unnorm p(z_K) \mathcal{N}\left(\rho_K \vert 2s\left(K\delta, z_K\right), I\right) \quad \mbox{and} \quad q(z_1, \rho_1) = q(z_1) \mathcal{N}\left(\rho_1 \vert 2 s\left(\delta, z_1\right), I\right),
\end{equation*}
and the result from \cref{lemma:ratio}, our framework yields
\begin{equation}
\frac{\unnorm p(\zg, \rhog)}{q(\zg, \rhog)} = \frac{\unnorm p(z_K) \mathcal{N}\left(\rho_K \vert 2s\left(K\delta, z_K\right), I\right)}{q(z_1) \mathcal{N}\left(\rho_1 \vert 2 s\left(\delta, z_1\right), I\right)} \prod_{k=1}^{K-1} \frac{\mathcal{N}(\rho_k \vert 2s(k\delta, z_k, I))}{\mathcal{N}(\rho_k'\vert 0, I)}.
\end{equation}

Finally, replacing $\rho_{k+1}$ and $\rho_k'$ by their expressions from \cref{eq:UHAforwardULA}, this ratio can be written as
\begin{equation}
\frac{\unnorm p(\zg, \rhog)}{q(\zg, \rhog)} = \frac{p(z_K)}{q(z_1)} \prod_{k=1}^{K-1} \frac{\mathcal{N}\left( \xi_k + \frac{\delta}{2} \nabla \log \pi_{k}(z_k) + \frac{\delta}{2} \nabla \log \pi_{k}(z_{k+1}) - 2s\left((k+1)\delta, z_{k+1}\right) \vert 0, I\right)}{\mathcal{N}(\xi_k \vert 0, I)},
\end{equation}
which recovers the ratio used by \MCD\ taking $\epsilon = \frac{\delta^2}{2}$ and $s\left((k+1)\delta, z_{k+1}\right) = -\frac{1}{2} \sqrt{2\epsilon} \, \tilde s\left(k\delta, z_{k+1}\right)$. This shows that our framework with the choices from \cref{thm:mcd} recovers \MCD.

\newpage

\subsection{Proof of \cref{thm:uha}}

For this proof we will use $\LP(z, \rho)$ to denote a single step of the leapfrog integrator typically used by HMC (see \cref{alg:Fldvi} for the definition), and $\gamma(z, \rho)$ to denote the operator that negates the momentum variables, that is, $(z, -\rho) = \gamma(z, \rho)$.

\UHA\ uses forward transitions $F_k(z_{k+1}, \rho_{k+1} \vert z_k, \rho_k)$ that consist of three steps
\begin{enumerate}
\item Resample momentum as $\rho_k' \sim \mathcal{N}(\eta \rho_k, (1 - \eta^2) I)$, where $\eta = \exp(-\gamma \delta)$,
\item Apply a leapfrog step $\LP$ followed by a negation of the momentum, which gives $(z_k'', \rho_k'') = (\gamma \circ \LP)(z_k, \rho_k')$,
\item Negate the momentum, which gives $(z_{k+1}, \rho_{k+1}) = \gamma(z_k'', \rho_k'')$.
\end{enumerate}
Combining steps (2) and (3) gives $(z_{k+1}, \rho_{k+1}) = (\gamma \circ \gamma \circ \LP)(z_k, \rho_k')$, which can be simplified to $(z_{k+1}, \rho_{k+1}) = \LP(z_k, \rho_k')$, since $\gamma$ is an involution (self-inverting). Thus, \UHA\ transitions can be expressed as a sequence of two steps: momentum resampling, followed by an application of a leapfrog step used by HMC. This is exactly the same as the forward transition from \cref{alg:Fldvi} with exact momentum resampling.

Similarly, the backward transitions $B_k(z_{k}, \rho_{k} \vert z_{k+1}, \rho_{k+1})$ used by \UHA\ also consist of three steps, given by
\begin{enumerate}
\item Negate the momentum, which gives $(z_{k}'', \rho_{k}'') = \gamma(z_{k+1}, \rho_{k+1})$.
\item Apply a leapfrog step $\LP$ followed by a negation of the momentum, which gives $(z_k, \rho_k') = (\gamma \circ \LP)(z_k'', \rho_k'')$,
\item Resample momentum as $\rho_k \sim \mathcal{N}(\eta \rho_k', (1 - \eta^2) I)$, where $\eta = \exp(-\gamma \delta)$.
\end{enumerate}
Using the fact that $\gamma \circ \LP = (\gamma \circ \LP)^{-1} = \LP^{-1} \circ \gamma^{-1}$ \citep{neal2011mcmc}, steps (1) and (2) above can be combined as $((\gamma \circ \LP) \circ \gamma)(z_{k+1}, \rho_{k+1}) = (\LP^{-1} \circ \gamma^{-1} \circ \gamma)(z_{k+1}, \rho_{k+1}) = \LP^{-1}(z_{k+1}, \rho_{k+1})$. This shows that the backward transitions used by \UHA\ can be expressed as a sequence of two steps: the inverse of a leapfrog step used by HMC, followed by exact resampling of the momentum. This is exactly the same as the backward transition from \cref{alg:Bldvi} with exact momentum resampling (possible due to removing the score network).

This shows that the forward and backward transitions used by \UHA\ are recovered by our framework with the simulation scheme from \cref{sec:discr}, using the choices stated in \cref{thm:uha}. Finally, using the momentum augmented distributions given by
\begin{equation*}
\unnorm p(z_K, \rho_K) = \unnorm p(z_K) \mathcal{N}(\rho_K \vert 0, I) \quad \mbox{and} \quad q(z_1, \rho_1) = q(z_1) \mathcal{N}(\rho_1 \vert 0, I),
\end{equation*}
and the result from \cref{lemma:ratio}, our framework yields the ratio
\begin{align*}
\frac{\unnorm p(\zg, \rhog)}{q(\zg, \rhog)} &= \frac{\unnorm p(z_K) \mathcal{N}(\rho_K \vert 0, I)}{q(z_1) \mathcal{N}(\rho_1\vert 0, I)} \prod_{k=1}^{K-1} \frac{\mathcal{N}(\rho_k \vert 0, I)}{\mathcal{N}(\rho_k'\vert 0, I)}\\
&= \frac{\unnorm p(z_K)}{q(z_1)} \prod_{k=1}^{K-1} \frac{\mathcal{N}(\rho_{k+1} \vert 0, I)}{\mathcal{N}(\rho_k'\vert 0, I)},
\end{align*}
which is exactly the ratio used by \UHA. This shows that our framework with the choices from \cref{thm:uha} recovers \UHA.

\newpage

\section{SIMPLER DISCRETIZATION SCHEME FROM SECTION \ref{sec:naive}} \label{app:dersimpler}

This section shows the derivation of the forward and backward transitions from \cref{alg:fsimple,,alg:bwsimple} together and an expression for their ratio, and results on all datasets using the resulting method.

\subsection{Transitions}

\paragraph{Forward transitions} The forward transitions $F_k^\mathrm{em}(z_{k+1}, \rho_{k+1} \vert z_k, \rho_k)$ from \cref{alg:fsimple} are obtained by splitting the forward SDE as
\begin{equation*}
\left[\begin{array}{c}
\dd z^t\\
\dd \rho^t
\end{array}\right] = 
\underbrace{\left[\begin{array}{c}
\rho^t \dd t \\
0
\end{array}\right]}_{\AN} + 
\underbrace{\left[\begin{array}{c}
0\\
\nabla \log \pi^t(z^t) \dd t - \gamma \rho^t \dd t + \sqrt{2\gamma} \dd w^t
\end{array}\right]}_{\BN},
\end{equation*}
and by sequentially composing the simulations for components $\BN\AN$. Component $\BN$ is simulated using the Euler-Maruyama scheme, while $\AN$ is simulated exactly. This yields the forward transitions from \cref{alg:fsimple}.

\paragraph{Backward transitions} The backward transitions $B_k^\mathrm{em}(z_k, \rho_k\ vert z_{k+1}, \rho_{k+1})$ from \cref{alg:bwsimple} are obtained by splitting the time-reversed SDE as
\begin{equation*}
\left[\begin{array}{c}
\dd y^t\\
\dd \lambda^t
\end{array}\right] = 
\underbrace{\left[\begin{array}{c}
-\lambda^t \dd t\\
0
\end{array}\right]}_{\ANB} + 
\underbrace{\left[\begin{array}{c}
0\\
-\nabla \log \pi^{T-t}(y^t) \dd t -\gamma \lambda^t \dd t + 2\gamma s(T-t, y^t, \lambda^t) \dd t + \sqrt{2\gamma} \dd w^t
\end{array}\right]}_{\BNB}
\end{equation*}
and by sequentially composing the simulations for components $\ANB\BNB$. Component $\ANB$ is simulated exactly, while $\BNB$ is simulated using the Euler-Maruyama scheme. This yields the backward transitions from \cref{alg:bwsimple}.

\paragraph{Ratio between transitions} The ratio between the transitions from \cref{alg:fsimple,,alg:bwsimple} is given by
\begin{equation} \label{eq:ratiosimpler}
\frac{B_k^\mathrm{em}(z_k, \rho_k\vert z_{k+1}, \rho_{k+1})}{F_k^\mathrm{em}(z_{k+1}, \rho_{k+1} \vert z_k, \rho_k)} = \frac{\mathcal{N}(\rho_k \vert \rho_{k+1}(1-\delta\gamma) - \delta \nabla \log \pi_{k\delta}(z_{k}) + 2\delta \gamma s(k\delta, z_k, \rho_{k+1}), 2\delta\gamma I)}{\mathcal{N}(\rho_{k+1} \vert \rho_k (1 - \gamma \delta) + \delta \nabla \log \pi_{k\delta}(z_k), 2\gamma\delta I)}.
\end{equation}
This can be obtained following a similar reasoning as the one used to prove \cref{lemma:ratio}.

\subsection{Results on all models} Results for all models are shown in \cref{tab:emion,,tab:emsonar,,tab:embrownian,,tab:emlorenz,,tab:emseeds}. In addition to \UHAEM\ and \LDVIEM, the tables include results for \ULA, \MCD, \UHA\ and \LDVI\ as well, to facilitate comparisons. It can be observed that, for all models, using the simpler transitions from \cref{alg:fsimple,,alg:bwsimple} (i.e. \UHAEM\ and \LDVIEM) lead to worse results than those obtained using the transitions from \cref{alg:Fldvi,,alg:Bldvi} (i.e. \UHA\ and \LDVI).

\begin{table}[]
\caption{ELBO achieved after training by different methods for different values of $K$ for a logistic regression model with the \textit{ionosphere} ($d=35$) dataset. Higher is better. Plain VI achieves an ELBO of $-124.1$ nats. Best result for each value of $K$ highlighted.}
\label{tab:emion}
\centering
\begin{tabular}{lcccccc}
\toprule
 & \multicolumn{6}{c}{Logistic regression (\textit{Ionosphere})} \\
\cmidrule(l{2pt}r{2pt}){2-7}
 & \ULA & \MCD & \UHA & \LDVI & \UHAEM & \LDVIEM\\
\midrule
$K=8$ & $-116.4$ & $-114.6$ & $-115.6$ & $\mathbf{-114.4}$ & $-117.7$ & $-115.5$ \\
$K=16$ & $-115.4$ & $-113.6$ & $-114.4$ & $\mathbf{-113.1}$ & $-115.9$ & $-113.8$ \\
$K=32$ & $-114.5$ & $-112.9$ & $-113.4$ & $\mathbf{-112.4}$ & $-114.6$ & $-112.9$ \\
$K=64$ & $-113.8$ & $-112.5$ & $-112.8$ & $\mathbf{-112.1}$ & $-113.6$ & $-112.4$ \\
$K=128$ & $-113.1$ & $-112.2$ & $-112.3$ & $\mathbf{-111.9}$ & $-113.1$ & $-112.1$ \\
$K=256$ & $-112.7$ & $-112.1$ & $-112.1$ & $\mathbf{-111.7}$ & $-112.5$ & $-111.9$ \\
\bottomrule
\end{tabular}
\end{table}

\begin{table}[]
\caption{ELBO achieved after training by different methods for different values of $K$ for a logistic regression model with the \textit{sonar} ($d=61$) dataset. Higher is better. Plain VI achieves an ELBO of $-138.6$ nats. Best result for each value of $K$ highlighted.}
\label{tab:emsonar}
\centering
\begin{tabular}{lcccccc}
\toprule
 & \multicolumn{6}{c}{{Logistic regression (\textit{Sonar})}} \\
\cmidrule(l{2pt}r{2pt}){2-7}
 & \ULA & \MCD & \UHA & \LDVI & \UHAEM & \LDVIEM \\
\midrule
$K=8$ & $-122.4$ & $-117.2$ & $-120.1$ & $\mathbf{-116.3}$ & $-124.1$ & $-118.5$ \\
$K=16$ & $-119.9$ & $-114.4$ & $-116.8$ & $\mathbf{-112.6}$ & $-119.9$ & $-114.4$ \\
$K=32$ & $-117.4$ & $-112.4$ & $-113.9$ & $\mathbf{-110.6}$ & $-116.4$ & $-111.7$ \\
$K=64$ & $-115.3$ & $-111.1$ & $-111.9$ & $\mathbf{-109.7}$ & $-113.8$ & $-110.3$ \\
$K=128$ & $-113.5$ & $-110.2$ & $-110.6$ & $\mathbf{-109.1}$ & $-111.9$ & $-109.6$ \\
$K=256$ & $-112.1$ & $-109.7$ & $-109.7$ & $\mathbf{-108.9}$ & $-110.7$ & $-109.1$ \\
\bottomrule
\end{tabular}
\end{table}

\begin{table}[]
\caption{ELBO achieved after training by different methods for different values of $K$ for the Brownian motion model ($d=32$). Higher is better. Plain VI achieves an ELBO of $-4.4$ nats. Best result for each value of $K$ highlighted.}
\label{tab:embrownian}
\centering
\begin{tabular}{lcccccc}
\toprule
 & \multicolumn{6}{c}{Brownian motion} \\
\cmidrule(l{2pt}r{2pt}){2-7}
 & \ULA & \MCD & \UHA & \LDVI & \UHAEM & \LDVIEM \\
\midrule
$K=8$ & $-1.9$ & $-1.4$ & $-1.6$ & $\mathbf{-1.1}$ & $-2.8$ & $-2.8$ \\
$K=16$ & $-1.5$ & $-0.8$ & $-1.1$ & $\mathbf{-0.5}$ & $-2.2$ & $-1.4$ \\
$K=32$ & $-1.1$ & $-0.4$ & $-0.5$ & $\mathbf{0.1}$ & $-1.6$ & $-0.5$ \\
$K=64$ & $-0.7$ & $-0.1$ & $0.1$ & $\mathbf{0.5}$ & $-0.9$ & $0.1$ \\
$K=128$ & $-0.3$ & $0.2$ & $0.4$ & $\mathbf{0.7}$ & $-0.4$ & $0.4$ \\
$K=256$ & $-0.1$ & $0.5$ & $0.6$ & $\mathbf{0.9}$ & $0.1$ & $0.6$ \\
\bottomrule
\end{tabular}
\end{table}

\begin{table}[]
\caption{ELBO achieved after training by different methods for different values of $K$ for the Lorenz system model ($d=90$). Higher is better. Plain VI achieves an ELBO of $-1187.8$ nats. Best result for each value of $K$ highlighted.}
\label{tab:emlorenz}
\centering
\begin{tabular}{lcccccc}
\toprule
& \multicolumn{6}{c}{Lorenz system} \\
\cmidrule(l{2pt}r{2pt}){2-7} 
 & \ULA & \MCD & \UHA & \LDVI & \UHAEM & \LDVIEM \\
\midrule
$K=8$ & $-1168.2$ & $-1168.1$ & $-1166.3$ & $\mathbf{-1166.1}$ & $-1170.5$ & $-1170.5$ \\
$K=16$ & $-1165.7$ & $-1165.6$ & $-1163.1$ & $\mathbf{-1162.2}$ & $-1169.8$ & $-1166.8$ \\
$K=32$ & $-1163.2$ & $-1163.3$ & $-1160.3$ & $\mathbf{-1157.6}$ & $-1167.9$ & $-1162.9$ \\
$K=64$ & $-1160.9$ & $-1161.1$ & $-1157.7$ & $\mathbf{-1153.7}$ & $-1161.3$ & $-1161.4$ \\
$K=128$ & $-1158.9$ & $-1158.9$ & $-1155.4$ & $\mathbf{-1153.1}$ & $-1158.1$ & $-1163.4$ \\
$K=256$ & $-1157.2$ & $-1157.1$ & $-1153.3$ & $\mathbf{-1151.1}$ & $-1163.1$ & $-1154.6$ \\
\bottomrule
\end{tabular}
\end{table}

\begin{table}[]
\caption{ELBO achieved after training by different methods for different values of $K$ for a random effect regression model with the \textit{seeds} dataset ($d=26$). Higher is better. Plain VI achieves an ELBO of $-77.1$ nats. Best result for each value of $K$ highlighted.}
\label{tab:emseeds}
\centering
\begin{tabular}{lcccccc}
\toprule
& \multicolumn{6}{c}{Random effect regression (seeds)} \\
\cmidrule(l{2pt}r{2pt}){2-7}
 & \ULA & \MCD & \UHA & \LDVI & \UHAEM & \LDVIEM \\
\midrule
$K=8$& $-75.5$ & $-75.1$ & $\mathbf{-74.9}$ & $\mathbf{-74.9}$ & $-75.9$ & $-75.5$ \\
$K=16$& $-75.2$ & $-74.6$ & $-74.6$ & $\mathbf{-74.5}$ & $-75.1$ & $-75.1$ \\
$K=32$& $-74.9$ & $-74.3$ & $\mathbf{-74.2}$ & $\mathbf{-74.2}$ & $-74.8$ & $-74.8$ \\
$K=64$& $-74.6$ & $-74.1$ & $-74.1$ & $\mathbf{-73.9}$ & $-74.4$ & $-74.4$ \\
$K=128$& $-74.3$ & $-73.9$ & $-73.8$ & $\mathbf{-73.7}$ & $-74.1$ & $-74.1$ \\
$K=256$& $-74.1$ & $-73.7$ & $-73.7$ & $\mathbf{-73.6}$ & $-73.9$ & $-73.7$ \\
\bottomrule
\end{tabular}
\end{table}

\end{document}